\documentclass[sigconf]{acmart}
\usepackage[utf8]{inputenc} % allow utf-8 input
\usepackage[T1]{fontenc}    % use 8-bit T1 fonts
\usepackage{url}            % simple URL typesetting
\usepackage{booktabs}       % professional-quality tables
\usepackage{amsfonts}       % blackboard math symbols
\usepackage{nicefrac}       % compact symbols for 1/2, etc.
\usepackage{microtype}      % microtypography
\usepackage{subcaption}
\usepackage{graphicx}
\usepackage[ruled,vlined,linesnumbered]{algorithm2e}
\usepackage{multirow}
\usepackage{amsmath}
\usepackage{url}
\usepackage{balance}
\usepackage{amsthm}
\usepackage{enumitem}
\usepackage{graphicx}
\include{pythonlisting}
\usepackage{graphics}
\newcommand*{\Scale}[2][4]{\scalebox{#1}{$#2$}}

\newtheorem{definition}{Definition}

\usepackage{xspace}

\DeclareMathOperator*{\argmax}{argmax}

\newcommand{\method}{\textsc{Eland}\xspace}
\newcommand{\methodi}{\textsc{Eland-itr}\xspace}
\newcommand{\methode}{\textsc{Eland-e2e}\xspace}
\newcommand{\gaug}{\textsc{GAug}\xspace}
\newcommand{\grand}{\textsc{Grand}\xspace}
\newcommand{\rnnfd}{\textsc{RNNfd}\xspace}
\newcommand{\jodie}{\textsc{JODIE}\xspace}
\newcommand{\dominant}{\textsc{Dominant}\xspace}
\newcommand{\deepae}{\textsc{DeepAE}\xspace}
\newcommand{\gcn}{\textsc{GCN}\xspace}
\newcommand{\gsage}{\textsc{GraphSAGE}\xspace}
\newcommand{\hetgnn}{\textsc{HetGNN}\xspace}
\newcommand{\evognn}{\textsc{EvolveGCN}\xspace}

\AtBeginDocument{%
  \providecommand\BibTeX{{%
    \normalfont B\kern-0.5em{\scshape i\kern-0.25em b}\kern-0.8em\TeX}}}

\copyrightyear{2021}
\acmYear{2021}
\setcopyright{acmcopyright}
\acmConference[CIKM '21]{Proceedings of the 30th ACM International Conference on Information and Knowledge Management}{November 1--5, 2021}{Virtual Event, Australia}
\acmBooktitle{Proceedings of the 30th ACM Int'l Conf. on Information and Knowledge Management (CIKM '21), November 1--5, 2021, Virtual Event, Australia}
\acmPrice{15.00}
\acmDOI{10.1145/3459637.3482313}
\acmISBN{978-1-4503-8446-9/21/11}

\makeatletter
\def\BState{\State\hskip-\ALG@thistlm}
\makeatother

\begin{document}
\fancyhead{}
\title[Action Sequence Augmentation for Early Graph-based Anomaly Detection]{Action Sequence Augmentation for \\ Early Graph-based Anomaly Detection}


\author{Tong Zhao*$^\dag$, Bo Ni*$^\dag$, Wenhao Yu$^\dag$, Zhichun Guo$^\dag$, Neil Shah$^\ddag$, Meng Jiang$^\dag$}
\affiliation{
	\institution{$\dag$ University of Notre Dame, IN, USA}
	\institution{$\ddag$ Snap Inc., WA, USA}
}
\email{{tzhao2, bni, wyu1, zguo5, mjiang2}@nd.edu,  nshah@snap.com}

\begin{abstract}
  The proliferation of web platforms has created incentives for online abuse. Many graph-based anomaly detection techniques are proposed to identify the suspicious accounts and behaviors. However, most of them detect the anomalies once the users have performed many such behaviors. Their performance is substantially hindered when the users' observed data is limited at an early stage, which needs to be improved to minimize financial loss. In this work, we propose \textsc{Eland}, a novel framework that uses action sequence augmentation for early anomaly detection. \textsc{Eland} utilizes a sequence predictor to predict next actions of every user and exploits the mutual enhancement between action sequence augmentation and user-action graph anomaly detection. Experiments on three real-world datasets show that \textsc{Eland} improves the performance of a variety of graph-based anomaly detection methods. With \textsc{Eland}, anomaly detection performance at an earlier stage is better than non-augmented methods that need significantly more observed data by up to 15\% on the Area under the ROC curve.
\end{abstract}

\thanks{* Equal contribution.}




\maketitle

\section{Introduction}
\label{introduction}
Social networks and review platforms indirectly create a market for malicious incentives, enabling malicious users to make huge profits via suspicious behaviors, e.g., fake reviews, hijacking trending topics. Such behaviors have severe negative impact on our society. User behavior data plays an essential role in the detection of malicious users. In the databases, each user creates a sequence of actions like giving a review to a particular item. In order to leverage the homophily of users, a great line of research work has been done to construct a user-item bipartite weighted graph and develop graph-based anomaly detection algorithms such as graph embeddings and graph neural networks. The weight is for the frequency of behaviors that associate the user and item in his/her action sequence. For example, \citet{kumar2018rev2} created a ``user-reviewed-product'' graph from each Amazon user's sequence of reviews; \citet{rayana2015collective} built ``user-reviewed-restaurants/hotels'' graphs from Yelp users' reviewing behaviors; \citet{zhao2020error} studied a ``user-posted-message'' graph from posting behaviors on social media.

\begin{figure}[t]
    \centering
    \includegraphics[width=0.8\linewidth]{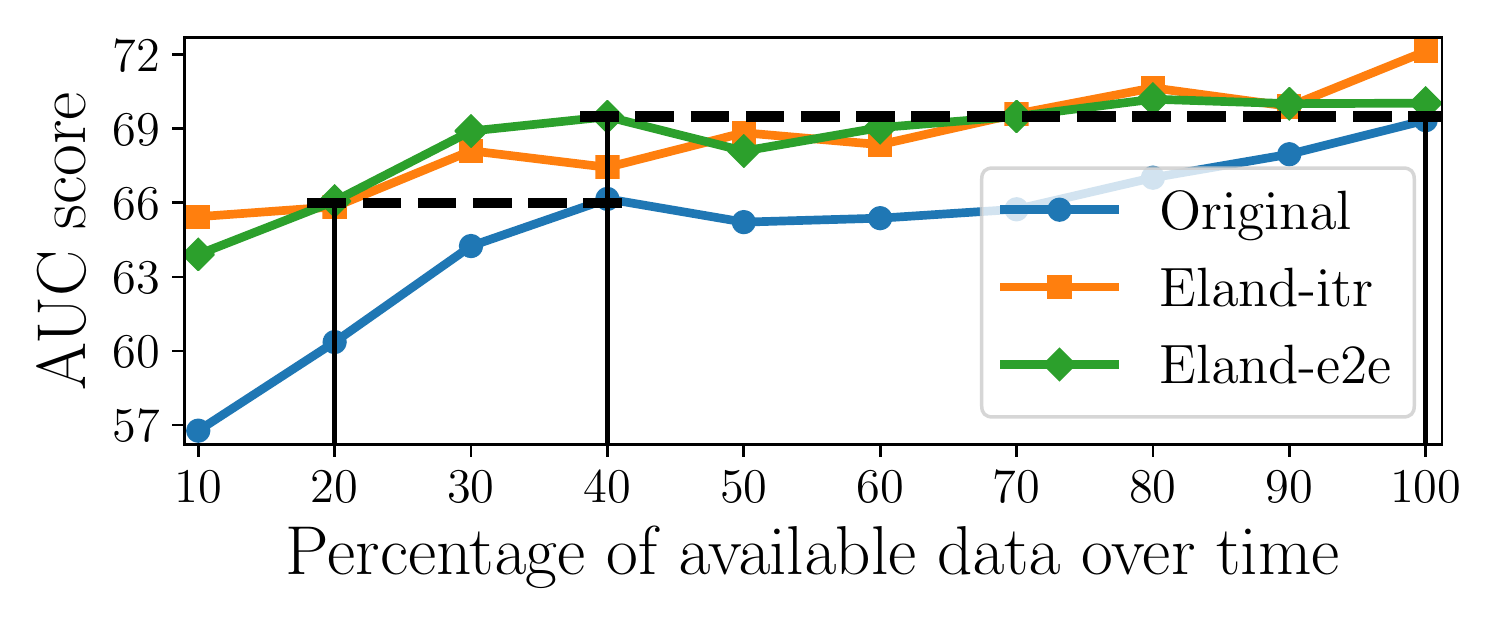}
\vspace{-0.15in}
\caption{Performance of \dominant~\cite{ding2019deep} and \method on a social media dataset considering the \emph{earliest} 10\%-100\% data from each user's action sequence. Both our \methodi and \methode with only 20\% (40\%) of available data can achieve the same performance as \dominant with 40\% (100\%) of data.}
\label{fig:dominant}
\vspace{-0.2in}
\end{figure}

Most of the existing work focuses on detecting anomalous users \emph{reactively}, i.e., when their malicious behaviors have already affected many people~\cite{shu2020leveraging}.
For example, a hijacked topic might have been on the trending list on Twitter for hours, and millions of users already saw and believed it; fake reviews on Yelp could have already damaged a restaurant's reputation.
Therefore, we argue that anomaly detection would be much more useful when it could be done \emph{early}, or \emph{proactively} to stop the malicious users before they achieve their targets.
In this work, we study the problem of \emph{early graph-based anomaly detection} when the observed behavior data is limited.

\begin{figure*}[t]
    \centering
    \includegraphics[width=0.95\textwidth]{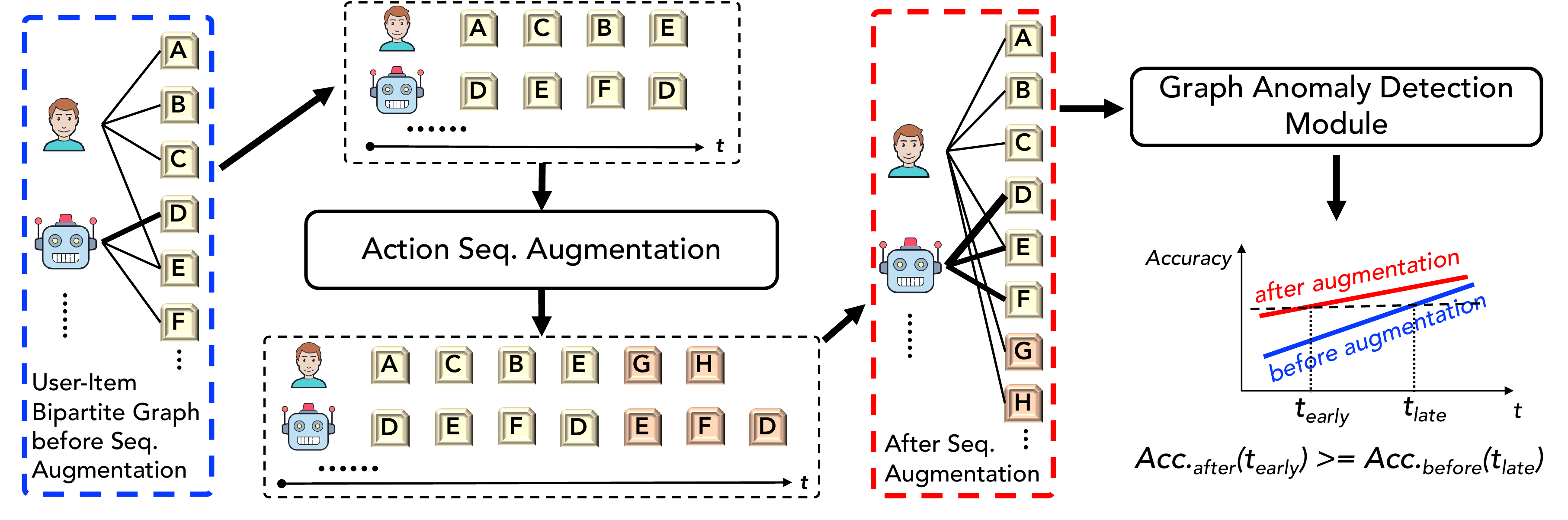}
    \vspace{-0.2in}
    \caption{\method framework: First, the dynamic
    \textcolor{blue}{original graph} is converted to action sequences, and then a sequence-based action predictor is trained to predict items that a user will adopt in the future based on his/her action history. 
    The predicted behaviors together with the \textcolor{blue}{original graph} form a richer, more informative \textcolor{red}{augmented graph}, which enables improvements upon a variety of graph anomaly detection methods.
    The significantly augmented data supports accurate early anomaly detection. \method improves the performance given equal observed data or comparable performance with less observed data. 
    }
    \label{fig:framework}
    \vspace{-0.1in}
\end{figure*}

Performing anomaly detection at an early stage is challenging due to the scarcity of available observations. Despite the technical advances of existing graph anomaly detection methods, we still witness substantially reduced performance in such settings where data is insufficient or incomplete. Empirically, we observe that state-of-the-art anomaly detection methods such as \dominant \cite{ding2019deep} would have a relative decrease of {15\%} on Area under the ROC curve (AUC) when only the earliest 20\% of the data (for each user) is available (see the blue curve in Figure~\ref{fig:dominant}).
Similar degradations occur on a few other types of graph learning methods such as \hetgnn~\cite{zhang2019heterogeneous} and \deepae~\cite{zhu2020anomaly} (Figure~\ref{fig:res_graph}),
leading us to ask: can we improve the performance at an early stage, when data is scarce?

Our idea is to learn and predict actions to augment the data at the early stage, and hence boost the performance of anomaly detection by ``forecasting the future.'' Although one anomalous user might not have sufficient behaviors to be detected, detection methods could still identify him with high confidence if his likely future behaviors are provided. That is, we predict the users' future behaviors by finding patterns from the entire data and prolonging  their action sequence with the items that they may adopt in the future.
With the predicted actions of a large number of users, the user-item (bipartite weighted) graph can be augmented to contain much richer information than before, thus enabling the detection methods to more accurately detect anomalies from the graph data.

\vspace{0.05in}
\noindent \textbf{Present work.} We propose \method (\underline{E}ar\underline{l}y \underline{An}omaly \underline{D}etection), a novel framework that achieves effective early graph anomaly detection via action sequence augmentation.
\method has two components: (1) a sequence augmentation module that predicts the user actions and augments the graph data and (2) an anomaly detection module that detects anomalous users from graph data. We present two methods to train the proposed framework: (1) \methodi where the two components are inter-dependent on each other and hence trained iteratively in a bootstrapping style, and (2) \methode where we jointly train the two modules as an end-to-end model. Shown in Figure \ref{fig:framework} is the illustration of the framework. The framework enables us to take advantage of the patterns of benign and malicious users discovered by the detection module to enhance the augmentation module, and vice versa.

Figure~\ref{fig:dominant} shows that (1) our \method (green and orange curves) that uses the users' earliest 20\% data can achieve the same performance as the original method (blue curve) that uses the earliest 40\%,  and (2) \method that uses 40\% data can achieve the performance of the original method with full data (100\%). 
Such observations indicate that \method could save \textbf{half the time} to collect users' data for accurate anomaly detection.

\vspace{0.05in}
The contributions of this work are summarized as follows: 
\begin{itemize}[leftmargin=1em,labelwidth=*,align=left] %nosep
\item We propose a novel idea that is to achieve early-stage graph anomaly detection by action sequence augmentation. Considering behavior data as sequences, we employ sequence prediction models (e.g., Seq2Seq) to forecast the behaviors.
\item We design a novel framework, \method, consisting of two components, action sequence augmentation and augmented graph anomaly detection to achieve early anomaly detection.
\item We conduct extensive experiments on three real-world datasets. \method achieves better performance on both unsupervised and supervised anomaly detection methods given less (earlier) data, with up to 15\% improvement on AUC score.
\end{itemize}

\section{Related work}
\label{related}
In this section we discuss three topics related to our work.

\vspace{0.05in}
\noindent \textbf{Graph Anomaly Detection} has received a great amount of academic interest in the past decade \cite{ding2019interactive,akoglu2015graph,zhao2018actionable,zhao2020error,zhao2021synergistic}.
Several methods \cite{kumar2018rev2,rayana2015collective,eswaran2018spotlight,ting2020isolation} were proposed following the graph outlier detection strategy. Similar approaches have been developed for bipartite graphs.
With the recent advances of graph neural networks (GNN), several GNN-based anomaly detection methods~\cite{ding2021few,wang2019semi,pang2019deep,dou2020enhancing,li2019specae} were proposed.
\textsc{Dominant}~\cite{ding2019deep} was an unsupervised attributed graph auto-encoder that detects anomalous nodes.
\citet{zhang2020repgcn} proposed multi-task GCN-based model that unified recommendation and anomaly detection.
\deepae~\cite{zhu2020anomaly} was an unsupervised graph auto-encoder-based method that detects anomalies by preserving multi-order proximity.
\citet{zhao2020error,zhao2021synergistic} proposed an unsupervised loss function that trains GNN to learn user representations tailored for anomaly detection.
% \citet{tian2020early} studied the early detection of rumors on Twitter.

\vspace{0.05in}
\noindent \textbf{Sequence Prediction.}
Sequential data prediction is a key problem in machine learning. For example, text generation aims to predict the next word on the given context~\cite{cho2014learning}; action prediction in computer vision aims to infer next action in video data~\cite{kong2017deep,liu2018ssnet}. Recurrent neural networks (RNN) have enjoyed considerable success in various sequence prediction tasks for understanding dynamical structure of data and producing accurate prediction.
Transformer models eschew recurrence and instead rely on the stacked multi-head self-attention to draw global dependencies between input tokens~\cite{vaswani2017attention}.
Recent work in recommender systems used sequence prediction to address the \emph{cold-start problem}~\cite{rama2019deep} besides many that created new graph neural models~\cite{liu2020heterogeneous,dong2020mamo} which ignore the sequential patterns in users' actions.
In our work, we focus on graph-based anomaly detection -- we predict the future actions to augment the user's action sequence, and exploit the mutual enhancement of sequence augmentation and graph anomaly detection.

\vspace{0.05in}
\noindent \textbf{Graph Neural Networks.}
As spectral GNNs generally operate on the full adjacency \cite{defferrard2016convolutional,kipf2016semi,ma2020unified}, spatial-based methods which perform graph convolution with neighborhood aggregation became prominent~\cite{hamilton2017inductive,velivckovic2017graph,gao2018large,guo2021few,zhao2021counterfactual},
owing to their scalability and flexibility~\cite{ying2018graph}. 
More recently, dynamic graph learning methods \cite{manessi2020dynamic,ma2020streaming} have proposed combining GNNs with RNNs to learn on dynamic graphs. GCRN \cite{seo2018structured} proposed a modified RNN by replacing fully connected layers with GCN layers \cite{kipf2016semi}. \evognn \cite{pareja2020evolvegcn} proposed to use GRU to learn the parameter changes in GCN instead of node representation changes.  \jodie~\cite{kumar2019predicting} proposed a user-item interaction prediction method based on historical interactions.  Such works have also been applied to large-scale industrial problems \cite{tang2020knowing,sankar2021graph}.

\section{Problem Definition}
\label{definition}
Consider a bipartite graph $\mathcal{G}=(\mathcal{U}, \mathcal{V}, \mathcal{E})$ at timestamp $t$, where $\mathcal{U}$ is the set of $m$ users, $\mathcal{V}$ is the set of $n$ items and $\mathcal{E}$ is the set of edges. 
Let $\mathbf{A} \in \mathbb{N}^{m\times n}$ be the adjacency matrix. Let $\mathbf{X}^u \in \mathbb{R}^{m\times k_u}$ and $\mathbf{X}^v \in \mathbb{R}^{n\times k_v}$ be the user feature matrix and item feature matrix, where $k_u$ and $k_v$ are the dimensions of raw features. As the feature dimensions of users and items are usually the same or can be projected to the same space via linear transformation, we use $\mathbf{X} \in \mathbb{R}^{(m+n)\times k}$ to denote the vertically concatenated feature matrix of $\mathbf{X}^u$ and $\mathbf{X}^v$.
Let the action sequence for each user $u$ be $x^{(u)}=\{\mathbf{x}^{(u)}_{1}, \dots, \mathbf{x}^{(u)}_{l_u}\}$, in which $l_u$ is the length of $u$'s action sequence and each item $\mathbf{x}^{(u)}_{i} \in \mathbb{R}^k$ stands for the feature vector of the corresponding item node. More actions between the same user-item pair result in with edges with higher weight.
We also denote $\mathbf{y} \in \{0, 1\}^m$ as labels for users where anomalies get $1$ and others get $0$. We follow the widely accepted definition of anomalous users in web graphs by previous works~\cite{jiang2016inferring,kumar2018rev2,zhao2020error,zhao2021synergistic}. For example, the anomalous user accounts in social networks are the botnets or the ones that frequently post advertisements or malicious links. 

Following the above notations and definitions, we define the task of early-stage graph-based anomaly detection.
Let $g: (\mathbf{A}, \mathbf{X}) \rightarrow \hat{\mathbf{y}}$ be a (supervised or unsupervised) graph anomaly detection method that returns a vector of prediction logits $\hat{\mathbf{y}} \in [0, 1]^m$.
Let $T$ be a later time when the observed data is ``sufficient'' for existing graph anomaly detection methods to perform well. We aim to design a framework that can achieve comparable or better performance at an earlier time $t < T$ when observations were ``incomplete.''\footnote{Although we consider the terms ``sufficient'' and ``incomplete'' relatively for discussion,  they can easily be made rigorous via practical constraints.}
Formally, our goal is to find a data augmentation framework satisfying the following criteria:
\begin{definition}{(Early-stage Graph Anomaly Detection)}
Let $h: (\hat{\mathbf{y}}\in\mathbb{R}^m, \mathbf{y}\in\mathbb{R}^m) \rightarrow \mathbb{R}$ be an evaluation metric (e.g., f-measure) such that a larger value is more desirable holding other conditions the same. Let $g$ be an anomaly detection method. Design an augmenter function $f: (\mathbf{A},\mathbf{X}) \rightarrow \mathbf{A}'$ that satisfies $h(g(f(\mathbf{A}, \mathbf{X}), \mathbf{X}),\mathbf{y}) \ge h(g(\mathbf{A}, \mathbf{X}), \mathbf{y})$. When assuming performance to increase monotonically with data, the above criteria is also equivalent to  $\exists T>t \mbox{ s.t. } $
\begin{equation}
h(g(f(\mathbf{A}, \mathbf{X}), \mathbf{X}), \mathbf{y}) \ge h(g(\mathbf{A}^*, \mathbf{X}), \mathbf{y})
\end{equation}
where $\mathbf{A}$ and $\mathbf{A}^*$ are the adjacency matrices at earlier time $t$ and later time $T$, respectively.
\end{definition}

When applying graph anomaly detection methods on real-world data, which usually has high complexity and uncertainty (and adversariality), it is difficult to theoretically guarantee that performance would monotonically increase with data.
In the following two sections, we first introduce our proposed framework \method which approximates $g(f(\mathbf{A}, \mathbf{X}), \mathbf{X})$, and show that \method \emph{empirically} satisfies the above desired property across choices of $g$, $\mathbf{A}$, and $\mathbf{X}$.

\section{The \method Frameworks}
\label{methods}
% The performance of existing methods are unsatisfactory for early anomaly detection due to the scarcity of available observations at an early time.
% Hence we propose to improve the performance of early anomaly detection via data augmentation, which is to provide the methods with more data with richer information.
% As the ``user-adopts-item'' bipartite graph (e.g., ``user-reviews-product'') is constructed from users' action sequences, we propose to adopt action sequence augmentation.
In this section, we first present the two major components of our proposed framework, \method. 
Then we introduce two ways of training \method: a bootstrapping-style iterative training \methodi that can be used on any existing graph anomaly detection methods;
and \methode that trains both modules together in an end-to-end fashion.

\subsection{Graph Anomaly Detection Module}
\label{sec:fd}
The first component of our proposed \method framework is a graph anomaly detection module ($g$). Notably, this part of \method is general and model-agnostic, in the sense that any anomaly detection or node classification model (e.g., \dominant\cite{ding2019deep}, \deepae\cite{zhu2020anomaly}, \gcn\cite{kipf2016semi}) suffices and can be used.  Without loss of generality, let the anomaly detection model $g_{ad}$ be defined as:
\begin{equation}
\label{eq:fd}
    \hat{\mathbf{y}} = g_{ad}( \mathbf{A}, \mathbf{X}; \Theta),
\end{equation}
where $\hat{\mathbf{y}} \in [0, 1]^m$ is the predicted suspiciousness of the user nodes of being anomalies, and $\Theta$ denotes trainable parameters. During training, if $g_{ad}$ is unsupervised method (e.g., \dominant\cite{ding2019deep}), its own training objective is used; if $g_{ad}$ is supervised method (e.g., \gcn\cite{kipf2016semi}), we use a standard binary cross-entropy loss.

Neural-based graph representation learning methods (e.g., graph neural networks) are capable of learning low-dimensional node representations as well as making predictions. We take advantage of the learned representations of user nodes. Without loss of generality, here we take the widely used Graph Convolutional Network (\gcn)~\cite{kipf2016semi} as an example of the anomaly detection method. The graph convolution operation of each \gcn layer is defined as:
\begin{equation}
    \mathbf{H}^{(j)} = \sigma(\tilde{\mathbf{D}}^{-\frac{1}{2}}\tilde{\mathbf{A}}\tilde{\mathbf{D}}^{-\frac{1}{2}}\mathbf{H}^{(j-1)}\mathbf{W}^{(j)}),
    \label{eq:gcn_layer}
\end{equation}
where $j$ indicates the layer, $\mathbf{H}^j$ is the node embedding matrix generated by $j$-th layer, $\mathbf{W}^{(j)}$ is the weight matrix of the $j$-th layer, $\tilde{\mathbf{A}} = \mathbf{A} + \mathbf{I}$ is the adjacency matrix with added self-loops,  $\tilde{\mathbf{D}}$ is the diagonal degree matrix  $\tilde{D}_{ii} = \sum_j \tilde{A}_{ij}$, and $\sigma(\cdot)$ denotes a nonlinear activation such as the Rectified Linear Unit (ReLU).

We write the GNN-based graph anomaly detection module as:
\begin{equation}
    \label{eq:fd-gnn}
    \hat{\mathbf{y}}, \mathbf{Z} = g_{ad-gnn}(\mathbf{A}, \mathbf{X}; \Theta),
\end{equation}
where $g_{ad-gnn}$ is a GNN-based model (e.g., \dominant~\cite{ding2019deep}, \gcn~\cite{kipf2016semi}), $\mathbf{Z}$ is the latent node representation matrix, and $\hat{\mathbf{y}}$ is the predicted user suspiciousness.

\subsection{Action Sequence Augmentation Module}
The action sequence augmentation module takes user action sequences as input and outputs the predicted next items that each user is likely to adopt in the future. These predicted actions are then added back to the ``user-adopts-item'' graph and forms the augmented graph. The main component of this module is a Seq2Seq encoder-decoder network~\cite{sutskever2014sequence}, which is robust to model choice and can be any state-of-the-art Seq2Seq model (e.g., GRU~\cite{cho2014learning}, Transformer~\cite{vaswani2017attention}, LSTM~\cite{hochreiter1997long}). 

The action sequence augmentation module is designed to capture behavior patterns from the sequential data (i.e., item adoption history) and use them for graph augmentation. In general, it takes the following functional form:
\begin{equation}
    \label{eq:gbf}
    \mathbf{A}' = f_{aug}(\mathbf{A}, \mathbf{X}, \hat{\mathbf{y}}; \Theta),
\end{equation}
where $\mathbf{A}'$ stands for the adjacency matrix with augmented predicted behaviors, $\mathbf{A}$ is the original adjacency matrix, $\mathbf{X}$ is the feature matrix, $\hat{\mathbf{y}}$ is the predicted user suspiciousness by $g_{ad}$ or $g_{ad-gnn}$, and $\Theta$ stands for the trainable parameter. 

\vspace{0.05in}
\noindent \textbf{Sequence prediction.} 
Here we regard each user's action sequence as a sequence of features that are constructed as prior knowledge. For example, for content-based item (e.g., reviews), the features can be the embedded representations of the texts. For each user $u$'s action sequence $x^{(u)}=\{\mathbf{x}^{(u)}_{1}, \dots, \mathbf{x}^{(u)}_{l_u}\}$, if a GNN-based anomaly detection module that learns node representation is used, say $\mathbf{z}_u$ for user $u$, then the action sequence is represented as $x^{(u)} = \{\mathbf{x}^{(u)}_{1} \oplus \mathbf{z}_u, \dots, \mathbf{x}^{(u)}_{l_u} \oplus \mathbf{z}_u\}$, where $\oplus$ stands for vector concatenation.

We opt for simplicity and adopt a GRU model to capture contextualized representations of each action in the sequence and make predictions. Hence, the hidden state of each action by user $u$ is:
\begin{equation}
    \label{eq:lstm}
    \mathbf{h}^{(u)}_i = \overrightarrow{\mathrm{GRU}}(\mathbf{x}^{(u)}_i, \mathbf{h}^{(u)}_{i-1}),
\end{equation}
where $\mathbf{h}_i$ refers to the hidden state in the $i$-th step.
Moreover, we use a linear readout function to predict of the next action by
\begin{equation}
    \hat{\mathbf{x}}^{(u)}_{i+1} = \mathbf{W}_p \cdot \mathbf{h}^{(u)}_i + \mathbf{b}_p,
    \label{eq:pred}
\end{equation}
where $\mathbf{W}_{p}\in \mathbb{R}^{|\mathbf{h}|\times k}$, $\mathbf{b}_{p}\in \mathbb{R}^{k}$ are trainable parameters and $\hat{\mathbf{x}}^{(u)}_{i+1}$ is the predicted feature vector for user $u$'s next action.  The decoder can also predict multiple sequential actions.

\vspace{0.05in}
\noindent \textbf{Sequence Augmentation.} Finally, we augment the graph by explicitly predicting the future items for users and adding the predicted behaviors as edges into the graph. 
When $\hat{\mathbf{x}}^{(u)}_{i+1}$ is predicted by Eq.(\ref{eq:pred}), the next input item is determined by cosine similarity:
\begin{equation}
    \label{eq:cos}
    \mathbf{x}^{(u)}_{i+1} = \argmax_{\mathbf{x}\in \mathbf{X}^v}\left(\frac{\hat{\mathbf{x}}^{(u)}_{i+1} \cdot \mathbf{x}}{||\hat{\mathbf{x}}^{(u)}_{i+1}|| \times ||\mathbf{x}||}\right).
\end{equation}
where $\mathbf{X}^v$ is the item feature matrix serving as the vocabulary. Let $v^{(u)}_{i}$ be the corresponding item to feature $\mathbf{x}^{(u)}_{i}$. For each user $u$, the decoder predicts the next $\Delta l_u$ future actions $\{v^{(u)}_{l_u+1}, ..., v^{(u)}_{l_u+\Delta l_u}\}$. 
Thereby, the augmented adjacency matrix can be calculated via
\begin{equation}
    \mathbf{A}' = \mathbf{A} + \sum_{u\in \mathcal{U}}\sum_{i = l_u+1} ^ {l_u+\Delta l_u} \mathbf{O}(u, v^{(u)}_i),
\end{equation}
where $\mathbf{O}$ is an empty matrix except $O_{i,j} = 1$.

We next discuss two strategies for training \method.

\begin{algorithm}[t]
\SetAlgoLined
\SetKwInOut{Input}{Input}
\SetKwInOut{Output}{Output}
\Input{Adjacency matrix $\mathbf{A}$; node feature matrix $\mathbf{X}$; number of iterations $I$; anomaly detection module $g \in \{g_{ad}, g_{ad-gnn}\}$; action sequence augmentation module $f_{aug}$.}
\Output{User prediction results $\hat{\mathbf{y}}$.}
$\mathbf{X}_{orig} = \mathbf{X}$ \;
\uIf {$g$ is $g_{ad-gnn}$} {
    $\hat{\mathbf{y}}, \mathbf{Z} = g(\mathbf{A}, \mathbf{X})$ \tcp*{Defined in Eq.(\ref{eq:fd-gnn})}
    $\mathbf{X} = \text{concat}(\mathbf{X}_{orig}, \mathbf{Z})$ \;
} 
\Else {
    $\hat{\mathbf{y}} = g(\mathbf{A}, \mathbf{X})$ \tcp*{Defined in Eq.(\ref{eq:fd})}
}
\For{i in range($I$)}{
    Re-initialize the parameters $\Theta$ in $f_{aug}$ \;
    $\mathbf{A}' = f_{aug}(\mathbf{A}, \mathbf{X}, \hat{\mathbf{y}})$  \tcp*{Defined in Eq.(\ref{eq:gbf})}
    \uIf {$g$ is $g_{ad-gnn}$} {
        Re-initialize the parameters $\Theta$ in $g$ \;
        $\hat{\mathbf{y}}, \mathbf{Z} = g(\mathbf{A}', \mathbf{X})$ \;
        $\mathbf{X} = \text{concat}(\mathbf{X}_{orig}, \mathbf{Z})$ \;
    }
    \Else {
        Re-initialize the parameters $\Theta$ in $g$ \;
        $\hat{\mathbf{y}} = g(\mathbf{A}', \mathbf{X})$ \;
    }
}
 return $\hat{\mathbf{y}}$ \;
 \caption{\methodi} 
 \label{alg:methodi}
\end{algorithm}

\subsection{\methodi: an Iterative Approach}
The graph anomaly detection module benefits from the enriched graph structure generated by the action sequence augmentation, and the augmentation module benefits from the detection module.  Both modules are interdependent and mutually enhance each other. Hence, we can use a bootstrapping training strategy to iteratively train both modules and jointly optimize their performances. 

Algorithm \ref{alg:methodi} shows the process of \methodi. We start with the anomaly detection module on the original graph. If the anomaly detection module is GNN-based, the features are updated by concatenating the learned node representations with the original node features. Then each iteration proceeds as follows: (1) we train a new action sequence augmentation module $f_{aug}$ with the graph and the results given by the detection module $g_{ad}$ or ${g_{ad-gnn}}$, (2) we train and make inference with a new initialized graph anomaly detection module on the updated graph structure $\mathbf{A}'$. After multiple iterations, the final prediction result $\hat{\mathbf{y}}$ is reported.

During the inference stage, we utilize the predicted suspiciousness scores $\hat{y}_u$ for each user given by the anomaly detection module to decide the number of predictions we make for user $u$:
\begin{equation}
    \label{eq:num_pred_i}
    \Delta l_u = \left\lfloor \kappa \cdot \hat{y}_u \right\rfloor,
\end{equation}
where $\kappa\in \mathbb{Z}^+$ is a hyperparameter to control the maximum number of predictions. The intuition is that anomalous users tend to perform more actions to achieve their goal (e.g., fake trending topic boosting), so we generate more predicted items for the users that are more likely to be anomalies, via the anomaly detection module.

\vspace{0.05in}
\noindent \textbf{Training \methodi.}
If $g_{ad}$ is unsupervised, it is trained with its own objectives. If supervised, the graph anomaly detection module is trained with the standard binary cross entropy:
\begin{equation}
    \label{eq:loss-fdgnn}
    \mathcal{L}_{ad} = - \sum_{u \in \mathcal{U}} \big( y_u\log(\hat{y}_u) + (1-y_u)\log(1-\hat{y}_u) \big).
\end{equation}

The action sequence augmentation module $f_{aug}$ is trained with the following loss function which maximizes the cosine similarity between the predicted actions and the actual actions. 
\begin{equation}
    \label{eq:loss-pred}
    \mathcal{L}_{aug} = -\frac{1}{|\mathcal{U}|} \sum_{u \in \mathcal{U}} \frac{1}{l_u}  \sum_{i = 1}^{l_u} \frac{\hat{\mathbf{x}}_i^{(u)} \cdot \mathbf{x}_i^{(u)}}{||{\mathbf{x}}_i^{(u)}|| \times ||\mathbf{x}_i^{(u)}||}.
\end{equation}

During the training of \methodi, the two modules ($g$ and $f$) are trained with the corresponding loss independently and iteratively.

\subsection{\methode: an End-to-End Approach}
In addition to \methodi, we propose an end-to-end model \methode that does not require the iterative training process. Hence, it avoids the potential error propagation issue in bootstrapping.

The anomaly detection module in \methode is a neural model that allows training with back-propagation in order to be trained together with the rest of the model. Thus, we use GNN-based models (e.g. \dominant~\cite{ding2019deep}) for the detection module. As \methode is end-to-end and trained as a whole and the action sequence augmentation module is executed prior to the graph anomaly detection module, the augmentation module should no longer require $\mathbf{y}'$ as input anymore. The module can then be defined as 
\begin{equation}
    \label{eq:gbf-e2e}
    \mathbf{A}' = f_{aug-e2e}(\mathbf{A}, \mathbf{X}; \Theta),
\end{equation}

When augmenting the graph with $f_{aug-e2e}$, instead of discretely predicting the next items as in Eq. (\ref{eq:cos}), we predict the following items via sampling. For each prediction of each user $u$, we use the cosine similarity of the predictions $\pi_{u} =[\pi_{u,1}, \dots, \pi_{u, n}]$ and apply the Gumbel-Softmax with reparameterization trick \cite{Jang2016gumbel,maddison2017concrete} to sample $\pi_{u}'$ from $\pi_{u}$ by:
\begin{equation}
    \label{eq:gumbel}
    \pi_{u}'=\frac{\exp \left(\left(\log \left(\pi_{u}\right)+g_{u}\right) / \tau\right)}{\sum_{j=1}^{n} \exp \left(\left(\log \left(\pi_{u,j}\right)+g_{j}\right) / \tau\right)},
\end{equation}
where $g_u{\sim}\mathrm{Gumbel}(0, 1)$ is a random variate sampled from the Gumbel distribution and $\tau$ is a temperature hyperparameter controlling the distribution of the results. Smaller $\tau$ results in more difference between classes. We then discretize $\pi_{u}'$ into one-hot vectors and add it into the adjacency matrix without damaging sparsity.

\begin{algorithm}[t]
\SetAlgoLined
\SetKwProg{Fn}{Function}{}{}
\SetKwInOut{Input}{Input}
\SetKwInOut{Output}{Output}
\Input{Adjacency matrix $\mathbf{A}$; node feature matrix $\mathbf{X}$; action sequence augmentation module $f_{aug-e2e}$; GNN-based anomaly detection module $g_{ad-gnn}$; number of training epochs $n\_epochs$.}
\Output{User prediction results $\hat{\mathbf{y}}$.}
\tcc{model training}
Initialize  $\Theta_{aug-e2e}$ in $f_{aug-e2e}$ and $\Theta_{ad-gnn}$ in $g_{ad-gnn}$ \;
\For{epoch in range(n\_epochs)}{
    $\mathbf{A}' = f_{aug-e2e}(\mathbf{A}, \mathbf{X})$  \tcp*{Defined in Eq.(\ref{eq:gbf-e2e})}
    $\hat{\mathbf{y}}, \mathbf{Z} = g_{ad-gnn}(\mathbf{A}', \mathbf{X})$ \tcp*{Defined in Eq.(\ref{eq:fd-gnn})}
    Calculate $\mathcal{L}_{e2e}$ with Eq.(\ref{eq:loss_e2e})\;
    Update $\Theta_{aug-e2e}$ and $\Theta_{ad-gnn}$ with $\mathcal{L}_{e2e}$ \;
}
\tcc{model inferencing}
$\mathbf{A}' = f_{aug-e2e}(\mathbf{A}, \mathbf{X})$ \;
$\hat{\mathbf{y}}, \mathbf{Z} = g_{ad-gnn}(\mathbf{A}', \mathbf{X})$ \;
return $\hat{\mathbf{y}}$ \;
\caption{\methode}
\label{alg:methode}
\end{algorithm}

As $f_{aug-e2e}$ no longer takes $\mathbf{y}'$ as input, we use preferential attachment to calculate the number of predictions for \methode. Zang et al. \cite{zang2019dynamical} showed that the dynamics of a random variable $x$ of exponential distribution $\frac{dx(t)}{dt}$ is proportional to $x(t)$ (i.e. $\frac{dx(t)}{dt} \propto x(t)$). Since user behavior patterns are generally considered to be consistent, it is reasonable to assume that the item selection process follows a Poisson process, whose frequency distribution could be further generalized to an exponential distribution~\cite{cooper2005poisson}.
Therefore, the number of predictions for user $u$'s actions can be calculated using preferential attachment:
\begin{equation}
    \Delta l_u = \left\lfloor \gamma \cdot \frac{d_u}{\sum d} \right\rfloor,
\end{equation}
where $d_u$ is $u$'s degree, $\sum d$ is the sum of all users' degree, and $\gamma$ is a hyperparameter that controls the total number of actions to augment the graph. $\Delta l_u$ actions will be predicted for $u$, taking the current distribution of actions into consideration. 

\vspace{0.05in}
\noindent \textbf{Training \methode.}
Depending on the graph anomaly detection module, its loss function $\mathcal{L}_{ad}$ can either be custom tailored (unsupervised) or a standard binary cross entropy loss (supervised) as defined in Eq.(\ref{eq:loss-fdgnn}). 
A cosine similarity loss $\mathcal{L}_{aug}$ as defined in Eq.(\ref{eq:loss-pred}) is still used to train the augmentation module.
Thus \methode is trained with a multi-task loss, defined as
\begin{equation}
    \label{eq:loss_e2e}
    \mathcal{L}_{e2e} = \mathcal{L}_{ad} + \mathcal{L}_{aug}.
\end{equation}
% where $\alpha$ is the weight of item-prediction loss.

Algorithm \ref{alg:methode} shows the process of \methode. In each epoch, the graph structure is augmented by the output from the action sequence augmentation module with the Gumbel-Softmax trick.  We use the Straight-Through gradient estimator~\cite{Jang2016gumbel}, passing gradients directly through un-discretized probabilities to train.  The GNN-based anomaly detection module uses the augmented graph to predict labels $\mathbf{y}'$. The multi-task loss $\mathcal{L}_{e2e}$ defined by Eq.(\ref{eq:loss_e2e}) is used to supervise the whole framework.

\begin{table}[t!]
% \small
% \scriptsize
  \caption{Statistics of the datasets.}
  \vspace{-0.1in}
  \label{tab:datasets}
  \centering
  \Scale[0.95]{\begin{tabular}{lrrr}
    \toprule
     & Weibo & Amazon & Reddit \\
    \midrule
    \# Users & 40,235 & 3,024 & 6,000 \\
    \# Items & 5,284 & 9,355 & 2,943 \\
    \# Edges & 55,624 & 2,434,019 & 79,210 \\
    \# Actions & 75,285 & 15,822,365 & 604,919 \\
    \% Anomaly users & 8.2\% & 20.2\% & 13.9\% \\
    \bottomrule
  \end{tabular}}
  \vspace{-0.05in}
\end{table}

\section{Experiments}
\label{experiments}

In this section, we evaluate the performance of proposed \methodi and \methode.  Our implementation is
made publicly available\footnote{\url{https://github.com/DM2-ND/Eland}}.

\subsection{Experimental Setup}

\subsubsection{Datasets} \hfill\\
We evaluate with three real-world datasets across different domains. 
\noindent \textbf{Weibo} is a micro-blogging dataset~\cite{jiang2016inferring}. We build ``user-(re)posted-microblog'' graph where the user nodes are registered users and item nodes are micro-blog posts. An action between a users and an item indicates that the user posted the item.
The original dataset~\cite{jiang2016inferring,zhao2020error} contains public user profiles, (re)posting of micro-blogs, and text of all micro-blogs. As the dataset is a micro-blogging graph, we define the anomalous users in this dataset as the social-spam users accounts which continuously post advertisements or malicious links~\cite{zhao2020error}. Due to the large scale (40K+ users), the dataset does not have manually annotated golden labels. So we labelled the users by their post text, profile information, and their behavior time following previous works~\cite{jiang2016inferring,zhao2020error}. Specifically, we use the following criteria to label anomalies: (1) \textit{Social spambot accounts:} The major conspicuous characteristic of suspicious users is their bot-controlled behavior. As we observed, most suspicious users posted in a fixed set of time intervals. For example, a user is identified as an anomaly if more than 2/3 of the time intervals of his posts are within $30 \pm 2$ seconds. (2) \textit{Accounts with suspicious posts:} We checked the post text and spotted the accounts whose posts are mostly advertisements or content with malicious links. 
% To further validate the above rule-based labels, we randomly sampled 5,000 user accounts with their post history for hand labeling by experts. Over 97\% of the rule-based labels and labels by two experts are consistent. 
We follow the train/validation/test split in previous works~\cite{zhao2020error} where we randomly pick 5000 users as training set, 5000 users as validation set, and the rest users as testing set.

\noindent \textbf{Amazon Reviews} is a review dataset from Amazon~\cite{mcauley2013amateurs}. We build ``user-used-word'' graph where user nodes are amazon users and item nodes are words with polarity bias from Hedonometer's word list~\cite{dodds2015human}. An action between a user and an item indicates that the user used that word in a review. 
The original dataset~\cite{mcauley2013amateurs} contains reviews on Amazon under the directory of Video Games. Following previous works~\cite{zhang2020repgcn,kumar2018rev2}, ground truth is is defined using helpfulness votes, which is indicative of malicious~\cite{kumar2018rev2} behavior. Users with at least 50 votes are labeled benign if the fraction of helpful-to-total votes is $\ge$ 0.75, and fraudulent if $\le$ 0.25. 
We randomly split the users in to train/validation/test sets with the ratio of 20\%/20\%/60\%.

\noindent \textbf{Reddit} is a forum dataset~\cite{baumgartner2020pushshift}. We build ``user-commented-subreddit'' graph where user nodes are reddit users and item nodes are subreddits. An action between a user and an item indicates that the user commented under that subreddit.
Due to the huge size of the original dataset~\cite{baumgartner2020pushshift} which contains entire public Reddit comments since 2005, we used the part of data from September 2019. As the dataset gives an score for each comment (\# of up votes minus \# of down votes), we label the users according similar rules for labeling of the Amazon dataset~\cite{zhang2020repgcn,kumar2018rev2}. We selected users who received at least 10 scores with absolute value $\ge 10$ (indicating at least 100 votes). Users are benign if all scores he/she received are positive; users are anomaly if $\ge 75\%$ of the scores he/she receive are $\le-10$.
We randomly split the users in to train/validation/test sets with the ratio of 20\%/20\%/60\%.

For all datasets, more actions between the same user and item would result in an edge with higher weight.
Statistics for the datasets are shown in Table~\ref{tab:datasets}. 

\begin{table*}[t]
  \caption{\method performance across supervised graph anomaly detection methods with only earliest 10\% of data available.}
  \label{tab:sup-10}
%   \small
  \vspace{-0.1in}
  \centering
  \Scale[0.95]{\begin{tabular}{cl|cc|cc|cc}
    \toprule
    Anomaly detection & Method & \multicolumn{2}{c}{Weibo} & \multicolumn{2}{|c}{Amazon Reviews} & \multicolumn{2}{|c}{Reddit} \\
    module $g_{ad}$ & & AUC & AP & AUC & AP & AUC & AP \\
    \midrule
    %%%%%%%%%% seq %%%%%%%%%%
    & \rnnfd~\cite{branco2020interleaved} & 54.52$\pm$0.12 & 17.44$\pm$0.10 & 60.22$\pm$0.29 & 28.61$\pm$0.11 & 66.08$\pm$0.36 & 26.45$\pm$0.83   \\
    & \grand~\cite{feng2020graph} & 82.58$\pm$2.11 & 40.12$\pm$2.99 & 81.71$\pm$2.56 & 57.66$\pm$2.98 & 79.09$\pm$0.18 & 42.37$\pm$0.72 \\
    \midrule
    %%%%%%%%%% GCN %%%%%%%%%%
    \multirow{5}*{\gcn \cite{kipf2016semi}}
    & Original & 81.78$\pm$0.78 & 41.21$\pm$1.36 & 80.28$\pm$0.09 & 57.73$\pm$0.21 & 78.01$\pm$0.71 & 41.21$\pm$0.69   \\
    & +\jodie~\cite{kumar2019predicting} & 67.80$\pm$1.30 & 17.12$\pm$2.72 & -- & -- & 73.12$\pm$2.13 & 31.62$\pm$3.98 \\
    & +\gaug~\cite{zhao2021data} & 82.04$\pm$0.40 & 48.17$\pm$0.59 & 81.91$\pm$0.02 & 60.12$\pm$0.15 & 78.78$\pm$0.07 & 40.74$\pm$0.72 \\
    & +\methodi & 82.76$\pm$0.71 & 48.51$\pm$1.06 & 80.85$\pm$0.67 & 58.14$\pm$0.39 & 78.94$\pm$0.83 &  43.11$\pm$1.22 \\
    & +\methode & \textbf{84.14}$\pm$0.50 & \textbf{54.15}$\pm$0.83 & \textbf{85.54}$\pm$0.46 & \textbf{65.48}$\pm$0.14 & \textbf{79.58}$\pm$0.38 & \textbf{44.60}$\pm$0.43 \\
    \midrule
    %%%%%%%%%% GSAGE %%%%%%%%%%
    \multirow{5}*{\gsage \cite{hamilton2017inductive}} 
    & Original & 81.87$\pm$0.56 & 45.26$\pm$2.54 & 78.67$\pm$0.09 & 58.00$\pm$0.07 & 81.06$\pm$0.02 & 47.71$\pm$0.01   \\
    & +\jodie~\cite{kumar2019predicting} & 69.44$\pm$0.95 & 16.01$\pm$2.09 & -- & -- & 74.66$\pm$0.09 & 34.70$\pm$0.06 \\
    & +\gaug~\cite{zhao2021data} & 82.10$\pm$0.46 & 47.81$\pm$1.29 & 80.79$\pm$0.02  & 56.38$\pm$0.03 & 81.37$\pm$0.01 & 43.83$\pm$0.01 \\
    & +\methodi & 82.34$\pm$0.50 & 48.40$\pm$0.91 & \textbf{81.59}$\pm$0.23 & \textbf{59.91}$\pm$0.12 & \textbf{81.62}$\pm$0.10 & \textbf{48.25}$\pm$0.11  \\
    & +\methode & \textbf{83.41}$\pm$0.37 & \textbf{50.61}$\pm$0.93 & 79.92$\pm$0.19 & 58.21$\pm$0.31 & 79.83$\pm$0.02 & 44.38$\pm$0.02  \\
    \midrule
    %%%%%%%%%% HETGNN %%%%%%%%%%
    \multirow{5}*{\hetgnn \cite{zhang2019heterogeneous}} 
    & Original & 81.33$\pm$0.43 & 39.66$\pm$1.48 & 86.24$\pm$0.13 & 67.98$\pm$0.25 & 91.51$\pm$0.13 & 67.51$\pm$0.17  \\
    & +\jodie~\cite{kumar2019predicting} & 68.99$\pm$0.44 & 17.38$\pm$1.87 & -- & -- & 92.02$\pm$0.36 & 68.16$\pm$0.30 \\
    & +\gaug~\cite{zhao2021data} & 82.09$\pm$0.21 & 47.05$\pm$0.51 & 87.26$\pm$0.12  & 71.76$\pm$0.33 & 91.99$\pm$0.02 & 66.30$\pm$0.25 \\
    & +\methodi & 81.46$\pm$0.57 & 40.20$\pm$1.19 & \textbf{90.58}$\pm$0.86 & \textbf{75.08}$\pm$0.57 & \textbf{92.44}$\pm$0.07 & \textbf{69.31}$\pm$0.29  \\
    & +\methode & \textbf{84.09}$\pm$0.55 & \textbf{54.07}$\pm$1.64 & 87.57$\pm$0.26 & 68.46$\pm$0.35 & 84.24$\pm$0.22 & 55.34$\pm$0.88  \\
    \bottomrule
  \end{tabular}}
\end{table*}

\begin{table*}[t]
  \caption{\method performance across unsupervised graph anomaly detection methods with only earliest 10\% of data available. \dominant and \deepae are not valid on Amazon data due to their design heuristic, hence their results are not included.}
  \label{tab:unsup-10}
%   \small
  \vspace{-0.1in}
  \centering
  \Scale[0.95]{\begin{tabular}{cl|cc|cc}
    \toprule
    Anomaly detection & Method &\multicolumn{2}{c}{Weibo} & \multicolumn{2}{|c}{Reddit} \\
    module $g_{ad}$ & & AUC & AP & AUC & AP \\
    \midrule
    %%%%%%%%%% DOMINANT %%%%%%%%%%
    \multirow{5}*{\shortstack{\dominant \cite{ding2019deep}\\(Unsupervised)}}
    & Original & 56.77$\pm$1.96 & 13.73$\pm$1.22 & 61.23$\pm$0.35 & 18.30$\pm$0.21   \\
    & +\jodie~\cite{kumar2019predicting} & 58.18$\pm$0.77 & 11.09$\pm$0.13 & 61.64$\pm$0.09 & 18.81$\pm$0.06 \\
    & +\gaug~\cite{zhao2021data} & 61.22$\pm$1.86 & 14.15$\pm$2.38 & 62.26$\pm$2.70 & 17.09$\pm$1.39  \\
    & +\methodi & \textbf{65.44}$\pm$1.78 & 19.42$\pm$1.29 & \textbf{62.96}$\pm$0.10 & 18.90$\pm$0.04  \\
    & +\methode & 63.91$\pm$0.92 & \textbf{21.90}$\pm$0.87 & 61.73$\pm$0.27 & \textbf{18.91}$\pm$0.14  \\
    \midrule
    %%%%%%%%%% DEEPAE %%%%%%%%%%
    \multirow{5}*{\shortstack{\deepae~\cite{zhu2020anomaly}\\(Unsupervised)}}
    & Original & 56.10$\pm$2.01 & 12.65$\pm$1.31 & 61.94$\pm$0.39 & 18.29$\pm$0.13   \\
    & +\jodie~\cite{kumar2019predicting} & 57.74$\pm$0.87 & 11.16$\pm$0.73 & 61.57$\pm$0.32 & 18.93$\pm$0.06 \\
    & +\gaug~\cite{zhao2021data} & 61.18$\pm$2.03 & 11.58$\pm$1.27 & 61.29$\pm$0.82 & 18.23$\pm$0.53  \\
    & +\methodi & \textbf{63.34}$\pm$0.82 & 15.88$\pm$0.73 & \textbf{62.87}$\pm$0.37 & \textbf{19.02}$\pm$0.11  \\
    & +\methode & 62.80$\pm$3.60 & \textbf{16.99}$\pm$3.87 & 62.47$\pm$0.11 & 18.88$\pm$0.04  \\
    \bottomrule
  \end{tabular}}
\end{table*}

\subsubsection{Baselines}\hfill\\
We evaluate \method with the following methods as $g_{ad}$ module:
\begin{itemize}[nosep,leftmargin=1em,labelwidth=*,align=left]
\item \gcn~\cite{kipf2016semi}: A supervised graph neural network. The neural structure has been empirically demonstrated to be useful in the area of graph anomaly detection \cite{zhang2020repgcn}. 
\item \gsage \cite{hamilton2017inductive}: An inductive graph neural network that can also be used for supervised node classification.
\item \hetgnn \cite{zhang2019heterogeneous}: A supervised graph neural network that handles the heterogeneous graphs of multiple types of nodes.
\item \dominant~\cite{ding2019deep}: An unsupervised graph anomaly detection method designed based on graph auto-encoder.
\item \deepae~\cite{zhu2020anomaly}: An unsupervised graph anomaly detection method with multi-order proximity preservation.
\end{itemize}
% We also investigate traditional graph mining methods for graph-based anomaly detection, \fraudar~\cite{hooi2016fraudar} and \lockinfer~\cite{jiang2016inferring}.
Moreover, we compare \method with the following baseline methods:
\begin{itemize}[nosep,leftmargin=1em,labelwidth=*,align=left]
\item \rnnfd~\cite{branco2020interleaved}: A RNN-based model for fraud detection on sequential data in industry applications.
\item \grand~\cite{feng2020graph}: A GNN-based model that leverages graph data augmentation to regularize the optimization process.
\item \jodie~\cite{kumar2019predicting}: A bipartite graph interaction prediction method by dynamic embedding trajectory learning.
\item \gaug~\cite{zhao2021data}: A graph data augmentation method designed for semi-supervised node classification with GNNs.
\end{itemize}
As \jodie~\cite{kumar2019predicting} and \gaug~\cite{zhao2021data} can be considered as model-agnostic graph data augmentation methods, we also report their performance across different $g_{ad}$ modules.

\subsubsection{Implementation Details}\hfill \\
All experiments were conducted on Linux servers with Intel Xeon Gold 6130 Processor (16 Cores @2.1Ghz), 96 GB of RAM, and 4 NVIDIA Tesla V100 cards (32 GB of RAM each) or RTX 2080Ti cards (11 GB of RAM each).

For all methods, we used hidden dimension of 128 and Adam optimizer. All methods have weight decay of $5\mathrm{e}{-4}$. For \gsage, we use the mean aggregator. To make fair comparisons, these aforementioned parameters are fixed for all experiments. For \gaug~\cite{zhao2021data}, we used the variant of \textsc{GAugM} because \textsc{GAugO} gets CUDA out of memory error on our datasets. We used the official implementation from the authors for \grand~\cite{feng2020graph}, \jodie~\cite{kumar2019predicting}, and \gaug~\cite{zhao2021data}.
We report the average and standard deviation of all performances in 20 runs with random parameter initialization.

\noindent \textbf{\methodi}: $\kappa$ is selected w.r.t. the average length of action sequences in each dataset. For Weibo and Reddit datasets, $\kappa = 150$; for Amazon Reviews dataset, $\kappa = 5000$.

\noindent \textbf{\methode}: As \methode is more robust to $\gamma$, we opt for simplicity and use $\gamma=100$ for all datasets. During the training of \methode, we linearly anneal the temperature of Gumbel-softmax distribution throughout the all training iterations, from $\tau = 5$ (a very flat distribution) to $\tau = 0.5$ (a very peaked distribution).

\subsection{Experimental Results}
% Table~\ref{tab:sup-10}, \ref{tab:unsup-10}, and \ref{tab:unsup-mining10} 
Table~\ref{tab:sup-10} and \ref{tab:unsup-10} 
report the performance of our proposed \method and baseline methods over supervised learning and unsupervised learning methods, respectively. These tables are organized per anomaly detection algorithm (row), per dataset (column), and per augmentation method (within-row). We report AUC and Average Precision (AP).
Note that results of \jodie~\cite{kumar2019predicting} on Amazon are missing due to CUDA out of memory when running the code from authors on V100 GPU with 32GB RAM.

\begin{figure*}[t]
    \centering
    \begin{subfigure}[b]{.245\linewidth}
        \includegraphics[width=\linewidth]{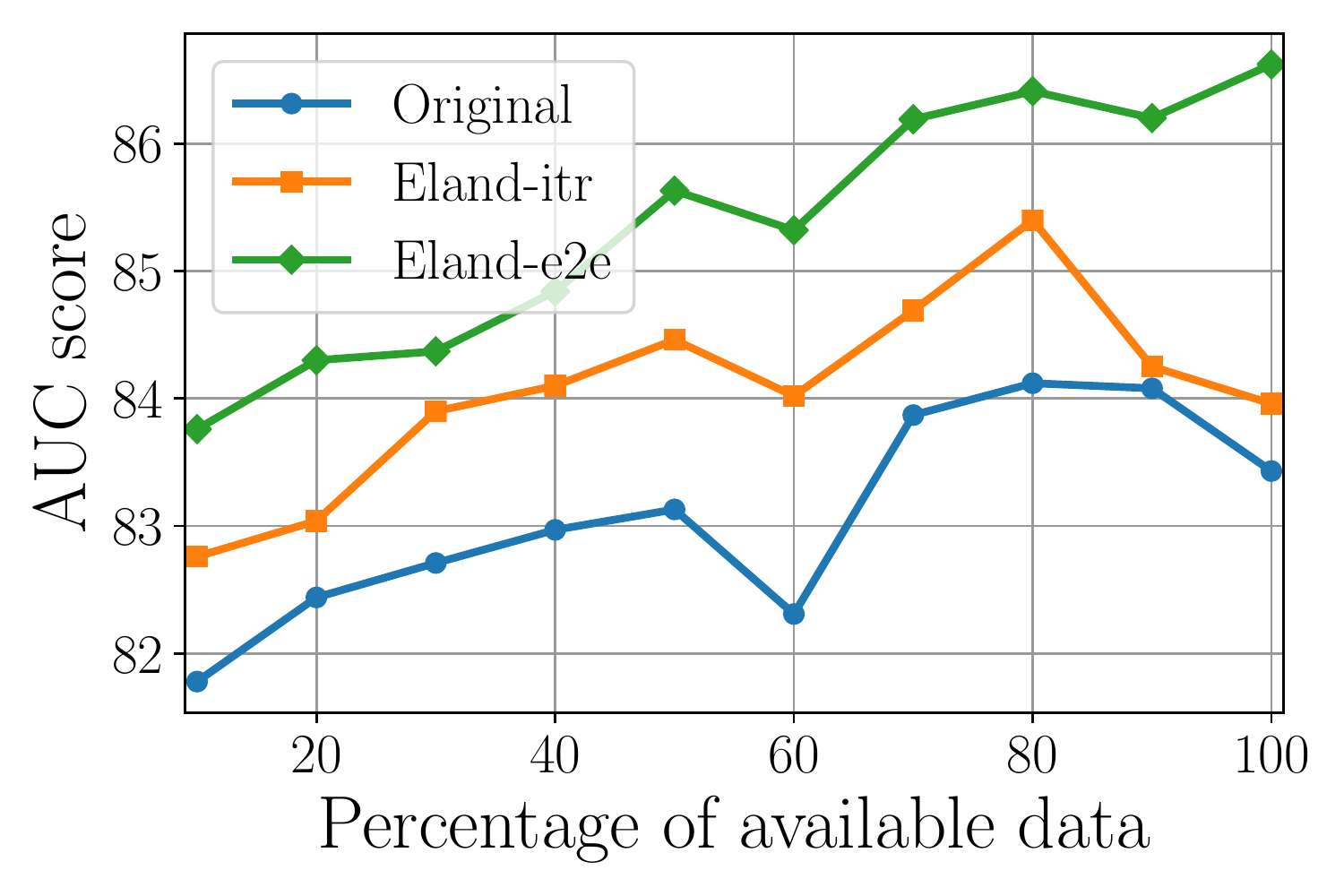}
        \caption{\gcn~\cite{kipf2016semi}}
    \end{subfigure}
    \begin{subfigure}[b]{.245\linewidth}
        \includegraphics[width=\linewidth]{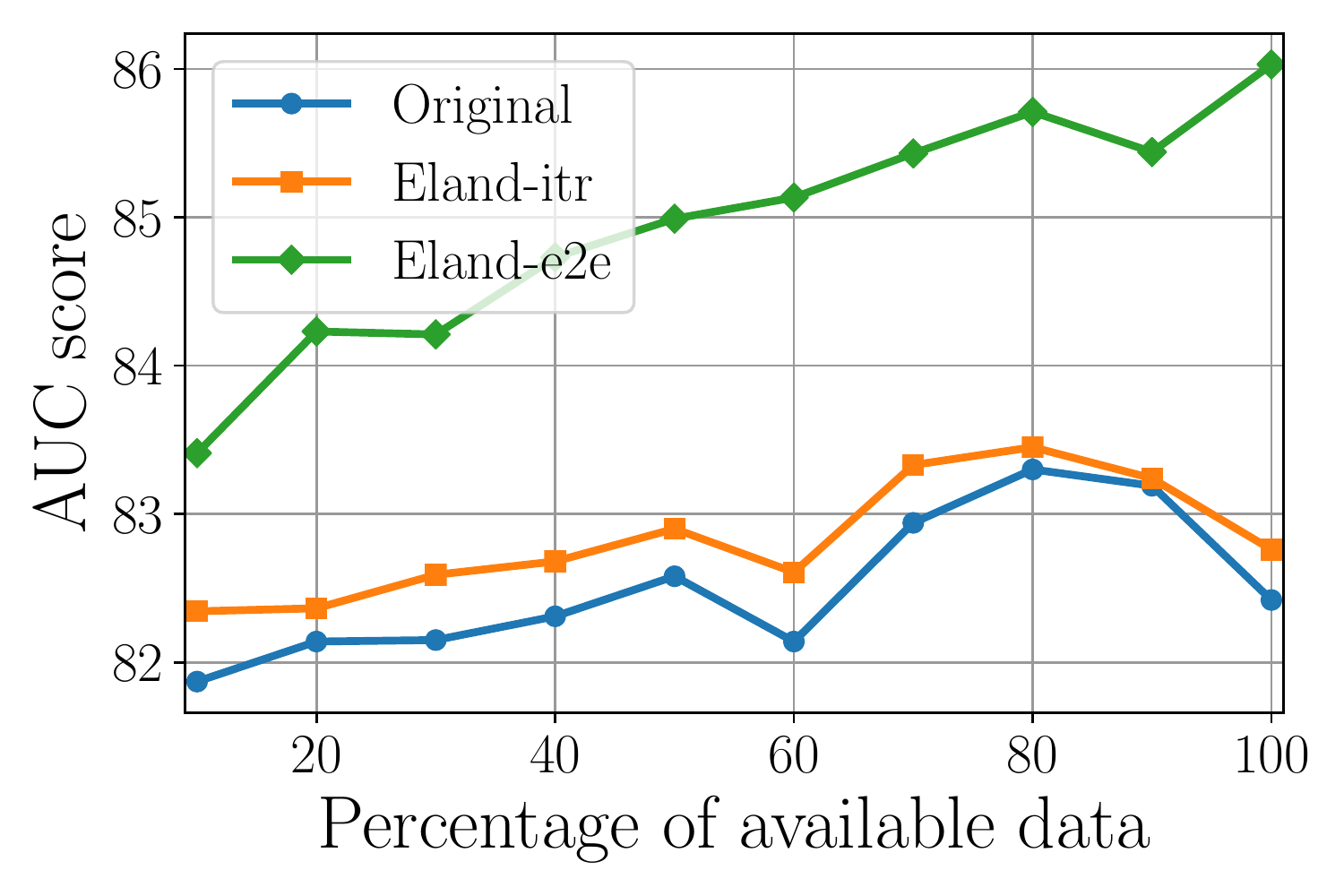}
        \caption{\gsage~\cite{hamilton2017inductive}}
    \end{subfigure}
    \begin{subfigure}[b]{.245\linewidth}
        \includegraphics[width=\linewidth]{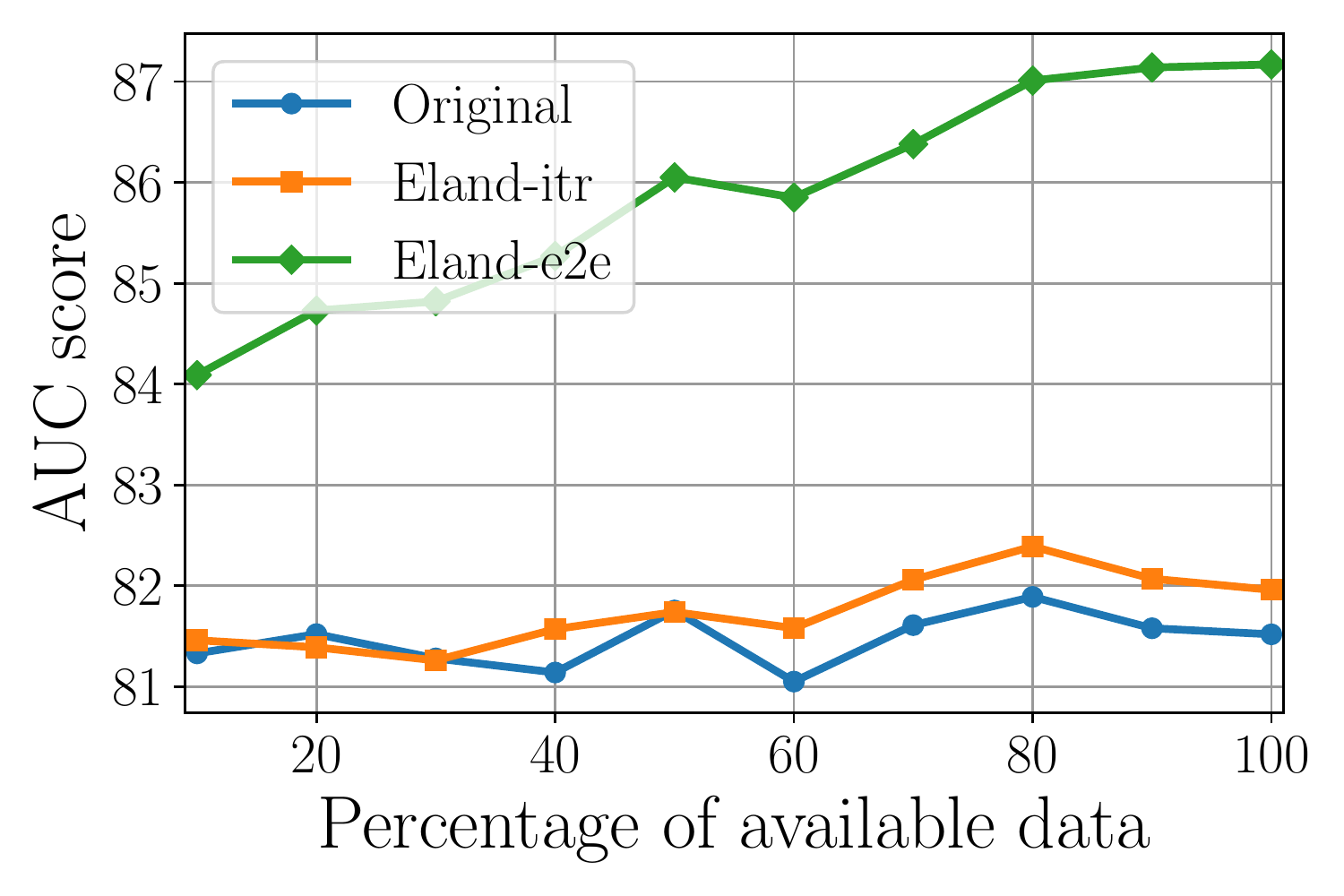}
        \caption{\hetgnn~\cite{zhang2019heterogeneous}}
    \end{subfigure}
    \begin{subfigure}[b]{.245\linewidth}
        \includegraphics[width=\linewidth]{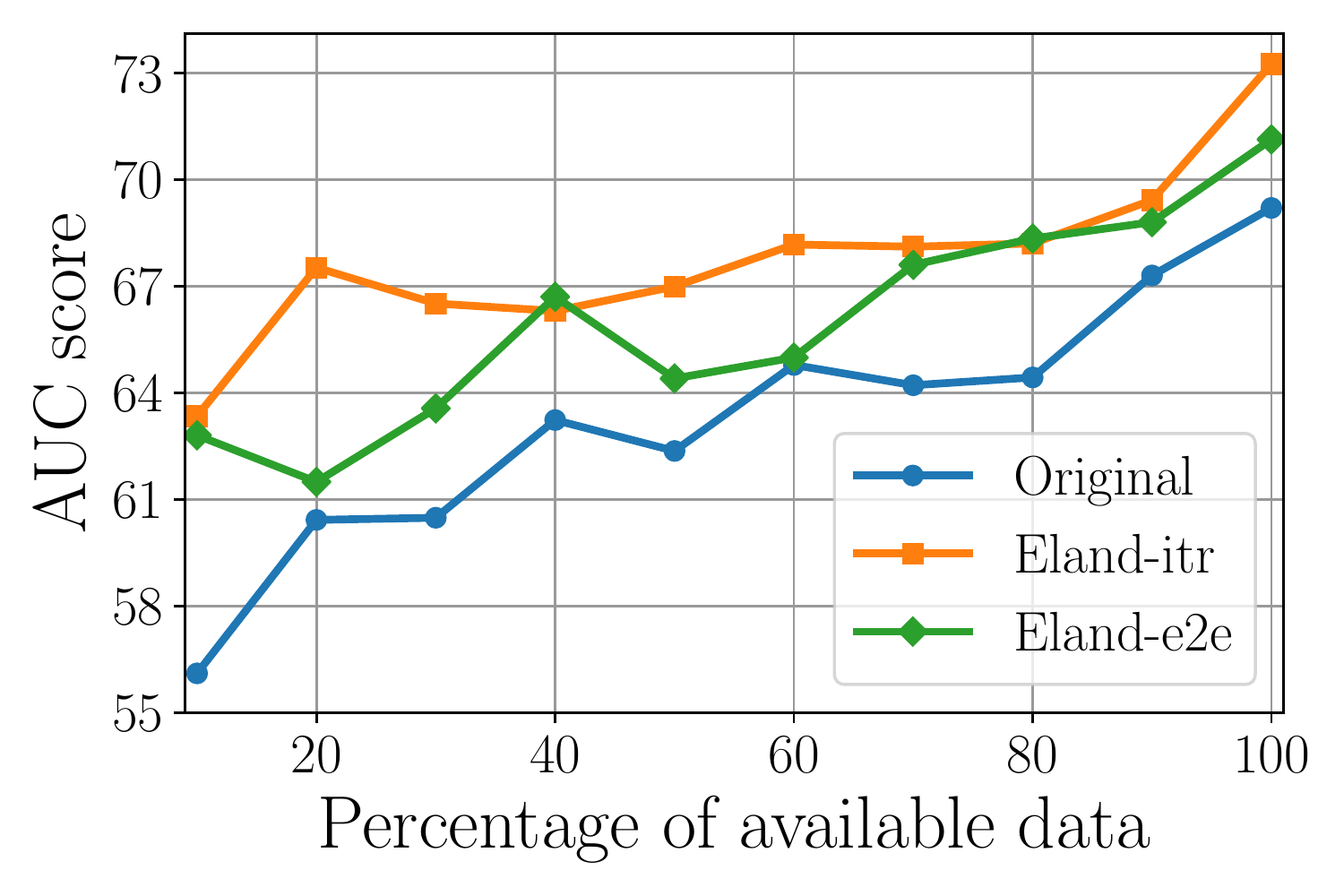}
        \caption{\deepae~\cite{zhu2020anomaly}}
    \end{subfigure}
\vspace{-0.in}
\caption{\methodi and \methode outperform the original version of (a)-(c) supervised graph learning methods and (d) unsupervised graph anomaly detection method in terms of AUC with different amount of available training data.}
\label{fig:res_graph}
\vspace{-0.in}
\end{figure*}

% \begin{table}[t]
%   \caption{\method performance across unsupervised graph mining-based anomaly detection methods on Weibo with only earliest 10\% of data available.}
%   \label{tab:unsup-mining10}
% %   \small
%   \vspace{-0.05in}
%   \centering
%   \Scale[0.95]{\begin{tabular}{cl|cc}
%     \toprule
%     Anomaly detection & Method & \multicolumn{2}{c}{Weibo} \\
%     module $g_{ad}$ & & AUC & AP  \\
%     \midrule
%     %%%%%%%%%% FRAUDAR %%%%%%%%%%
%     \multirow{2}*{\fraudar \cite{hooi2016fraudar}} 
%     & Original & 51.46 & 13.69  \\
%     & +\methodi & \textbf{51.98}$\pm$0.12 & \textbf{14.09}$\pm$0.03  \\
%     \midrule
%     %%%%%%%%%% LOCKINFER %%%%%%%%%%
%     \multirow{2}*{\lockinfer \cite{jiang2016inferring}} 
%     & Original & 53.69 & 8.43  \\
%     & +\methodi & \textbf{60.71}$\pm$0.58 & \textbf{15.96}$\pm$0.26   \\
%     \bottomrule
%   \end{tabular}}
% \end{table}

\subsubsection{Enhancing graph anomaly detection methods}\hfill\\
Table~\ref{tab:sup-10} shows that \method achieves improvements over baseline supervised graph anomaly detection methods, with the only exception of \methode with \gsage and \hetgnn on Reddit. Specifically, \methodi improves (averaged across datasets) 1.0\% (\gcn), 1.7\% (\gsage), and 2.1\% (\hetgnn) on AUC score and 7.7\%, 3.8\%, and 4.8\% on AP; \methode improves 3.8\%, 0.7\%, and 3.1\% on AUC and 17.7\%, 1.7\%, and 6.3\% on AP, respectively. 

We also observe that \method outperforms all four baseline methods. For \rnnfd, its results are not as good as any other methods as it only used the action sequence data.
For \grand, our proposed \methodi and \methode achieve 3.1\% and 4.1\% improvements on AUC, respectively.
\jodie hurts the performances of the original GNNs possibly because it was designed for predicting the next interaction item but not the following action sequence. Hence \method outperforms \jodie with large margins. On average, \methodi and \methode improve 12.7\% and 14.7\% on AUC, respectively.
Finally, \method also outperforms the graph data augmentation method \gaug. On average, \methodi and \methode improve 5.4\% and 1.1\% on AUC, respectively. 
We observe that \methode tend to perform better for supervised graph anomaly detection methods.

From Table~\ref{tab:unsup-10} we observe that \method outperforms unsupervised graph anomaly detection baselines. Specifically, \methodi improves by 9.0\% (\dominant), and 7.2\% (\deepae) on AUC and 22.4\%, and 14.8\% on AP; \methode improves 6.7\%, and 6.4\% on AUC and 31.4\%, and 18.8\% on AP, respectively. 
\method also outperforms both alternative methods \jodie and \gaug. On average, \methodi improves 5.1\% on AUC and 25.9\% on AP; \methode improves 3.6\% on AUC and 33.2\% on AP.
We note that \methodi achieves better performances than \methode with unsupervised anomaly detection modules. 
This is due to the misalignment of the training objectives of unsupervised methods and the evaluation metrics, resulting in the unsupervised objectives that mislead the action sequence augmentation module during end-to-end training.

\begin{table}[t]
\caption{\method is robust to the choice of Seq2Seq model used in the action sequence augmentation module. }
\label{tab:ablation}
% \small
\vspace{-0.in}
\Scale[0.95]{\begin{tabular}{cl|cc}
\toprule
Metric & Method & Weibo 10\% & Weibo 20\% \\
\midrule
\multirow{4}*{AUC} & \gcn & 81.78$\pm$0.78 & 82.44$\pm$0.54 \\
& +\methode (LSTM) & 83.76$\pm$0.74 & 84.30$\pm$0.53 \\
& +\methode (RNN) & 83.83$\pm$0.34 & 84.33$\pm$0.30 \\
& +\methode (GRU) & 84.14$\pm$0.50 & 84.57$\pm$0.36 \\
\midrule
\multirow{4}*{AP} & \gcn & 41.21$\pm$1.36 & 44.64$\pm$3.25 \\
& +\methode (LSTM) & 51.31$\pm$2.42 & 51.46$\pm$1.12 \\
& +\methode (RNN) & 52.74$\pm$0.63 & 52.84$\pm$0.92 \\
& +\methode (GRU) & 54.15$\pm$0.83 & 55.14$\pm$0.66 \\  
\bottomrule
\end{tabular}}
% \vspace{-0.1in}
\end{table}

We find that on the Amazon dataset, \dominant and \deepae are not effective, so their results are not included. The reason is that the unsupervised training objectives in the two methods are based on outlier detection with auto-encoders, which is not aligned with the actual label distributions. Therefore, most users that were marked as ``highly suspicious'' by these two methods in the Amazon data are actually benign users.

% Table~\ref{tab:unsup-mining10} shows the performance of \methodi with traditional unsupervised graph mining-based anomaly detection methods \fraudar~\cite{hooi2016fraudar} and \lockinfer~\cite{jiang2016inferring}, which are not neural based and hence cannot be trained end-to-end. We observe that \methodi is also capable of improving their performance.

\label{sec:table-results}

\subsubsection{Achieving early anomaly detection}\hfill\\
To better show the results of the proposed \method framework for early anomaly detection, we present Figure \ref{fig:res_graph} (along with Figure \ref{fig:dominant} for \dominant), in which we show the trends of AUC scores of original graph anomaly detection methods and our proposed framework change as more observed data is available on the Weibo dataset. 
We observe that in general, both \methodi and \methode are able to accomplish the early-stage anomaly detection task as defined in Section \ref{definition}. \methode generally achieves more improvements on supervised graph anomaly detection methods; for unsupervised methods, \methodi tends to perform better, which aligns with our observations from Table~\ref{tab:sup-10} and~\ref{tab:unsup-10}.

\begin{figure}[t]
% \vspace{-0.1in}
    \centering
    \begin{subfigure}[b]{.49\linewidth}
        \includegraphics[width=\linewidth]{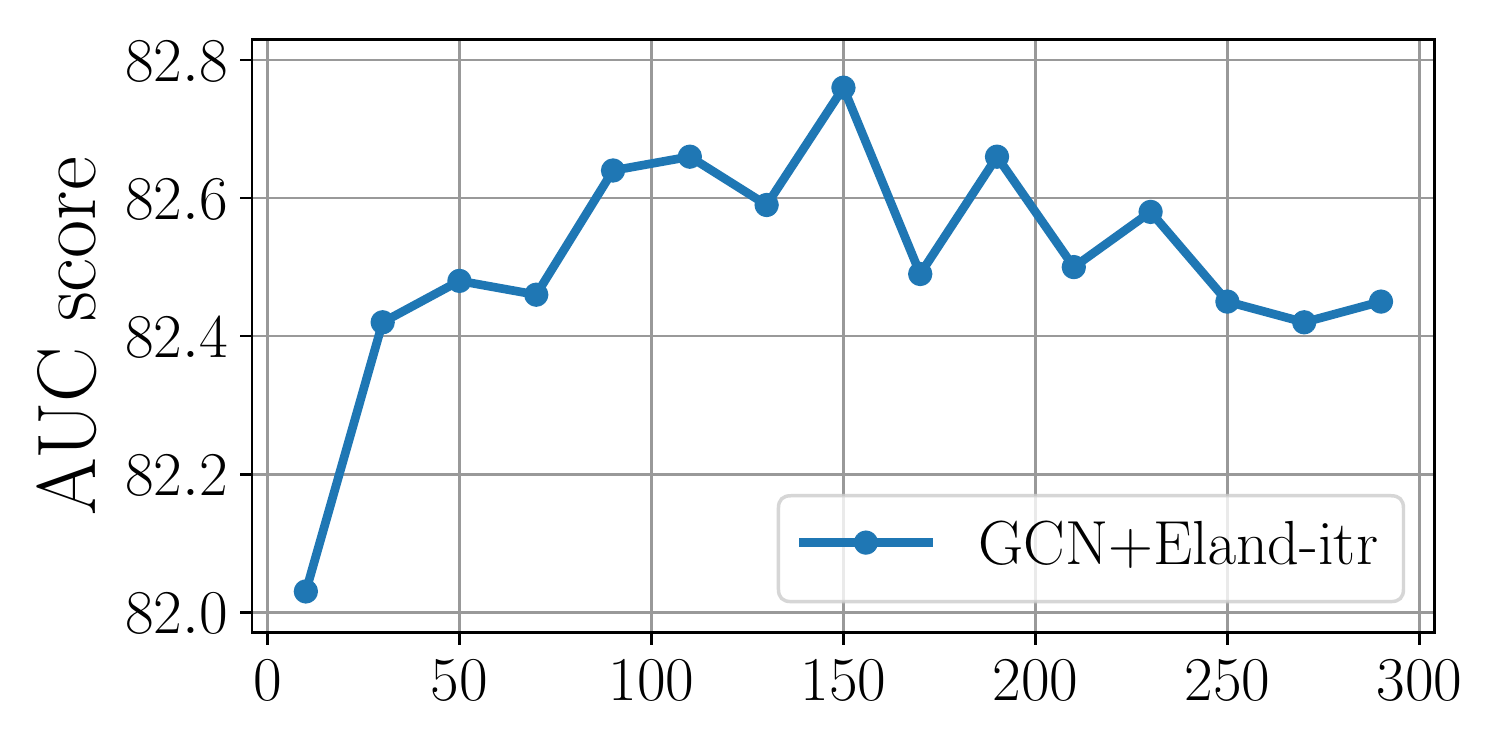}
        \vspace{-0.23in}
        \caption{$\kappa$}
        \label{fig:kappa}
    \end{subfigure}
    \begin{subfigure}[b]{.49\linewidth}
        \includegraphics[width=\linewidth]{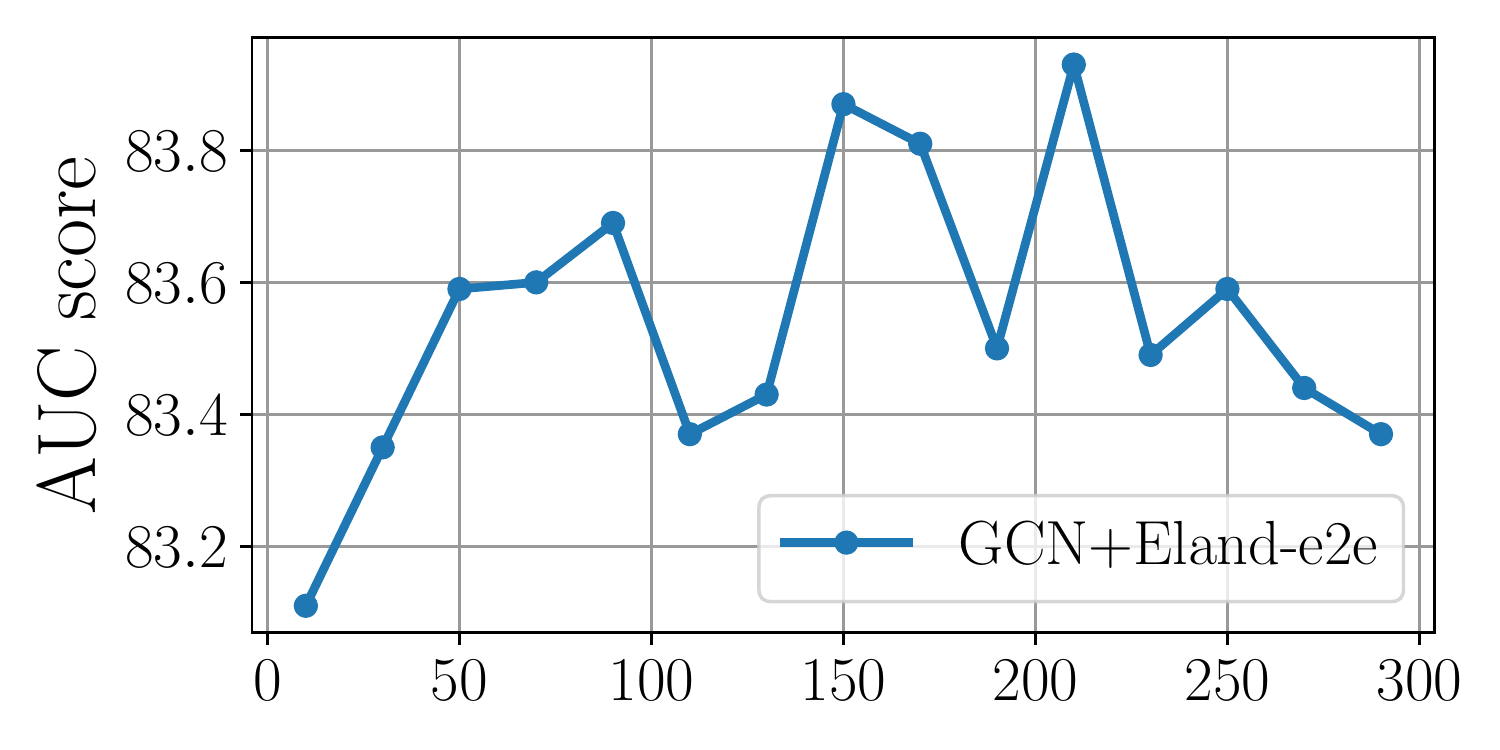}
        \vspace{-0.23in}
        \caption{$\gamma$}
        \label{fig:gamma}
    \end{subfigure}
    \vspace{-0.1in}
    \caption{\methodi and \methode are robust to the hyperparameters $\kappa$ and $\gamma$.}
  \label{fig:sens}
%   \vspace{-0.05in}
\end{figure}

\begin{figure}[t]
% \vspace{-0.1in}
    \centering
    \begin{subfigure}[b]{.49\linewidth}
        \includegraphics[width=\linewidth]{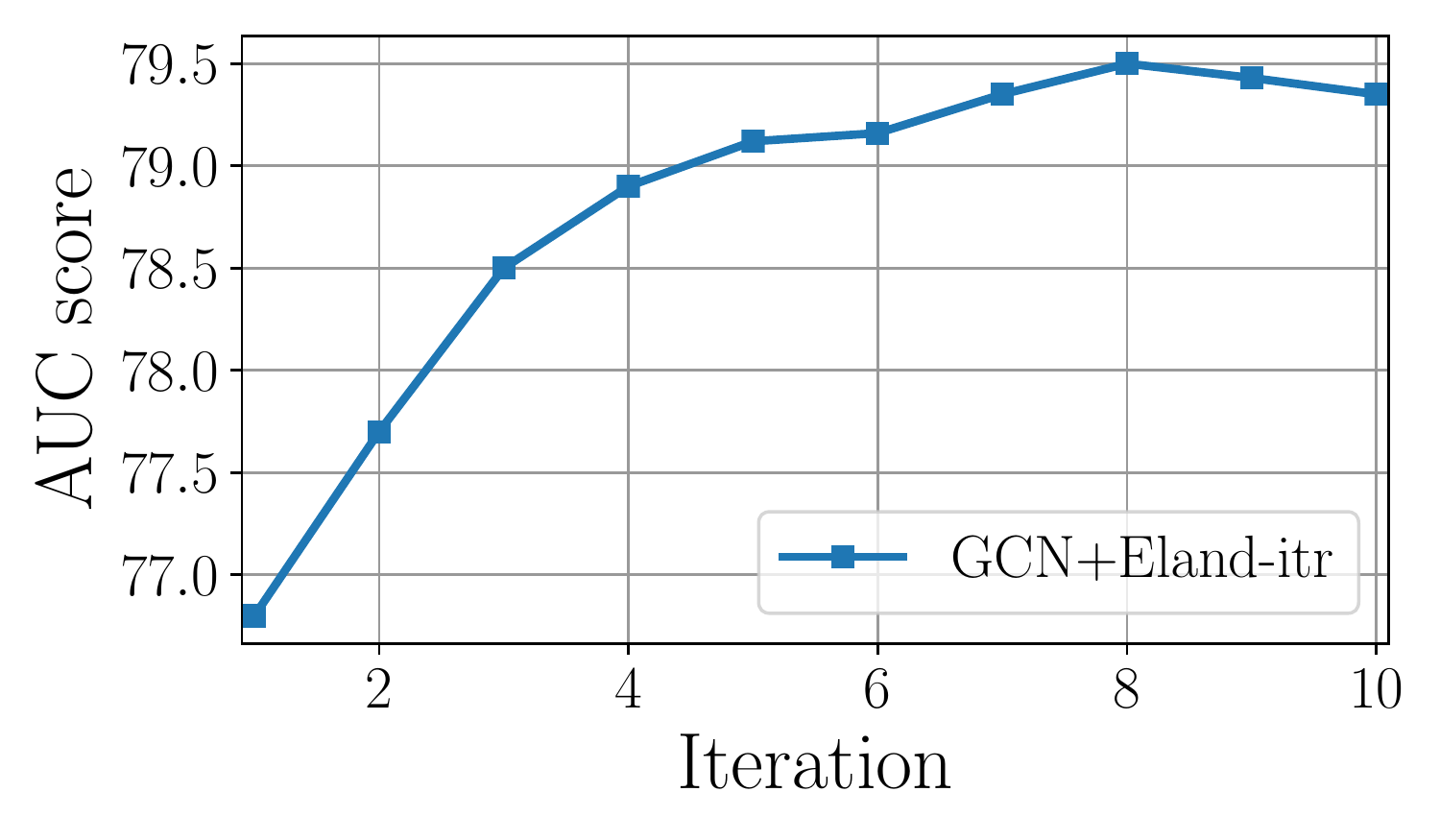}
        \caption{\methodi with \gcn}
        \label{fig:conv1}
    \end{subfigure}
    % \qquad \qquad
    \begin{subfigure}[b]{.49\linewidth}
        \includegraphics[width=\linewidth]{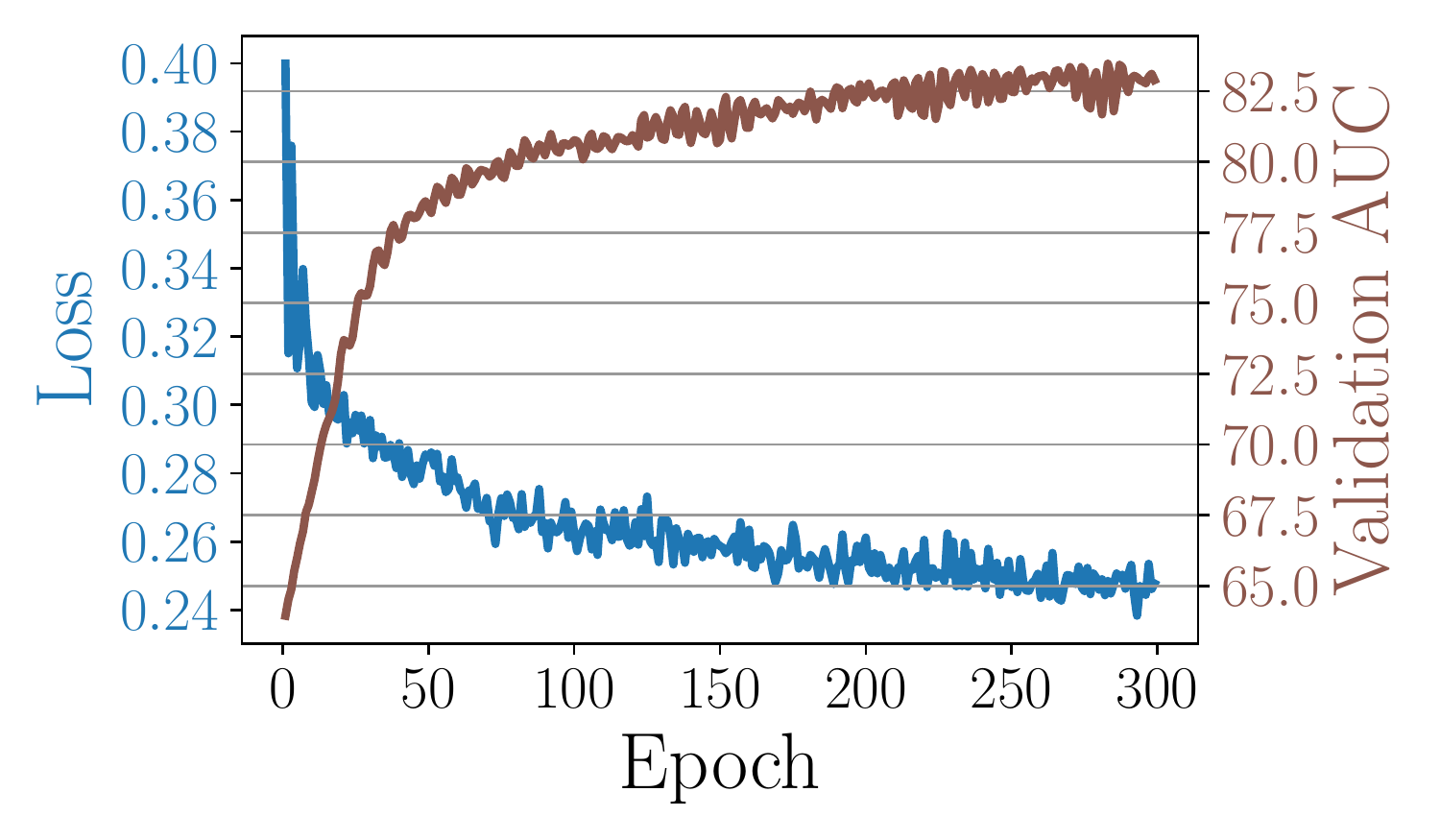}
        \caption{\methode with \gcn}
        \label{fig:conv2}
    \end{subfigure}
    \vspace{-0.05in}
    \caption{Convergence process of the proposed \methodi and \methode. When it came to the 8th iteration (in \methodi) or about 200 epochs (in \methode), the AUC of the two methods converge.}
  \label{fig:conv}
%   \vspace{-0.05in}
\end{figure}

\begin{figure*}[t]
% \vspace{-0.1in}
\centering
\includegraphics[width=\linewidth]{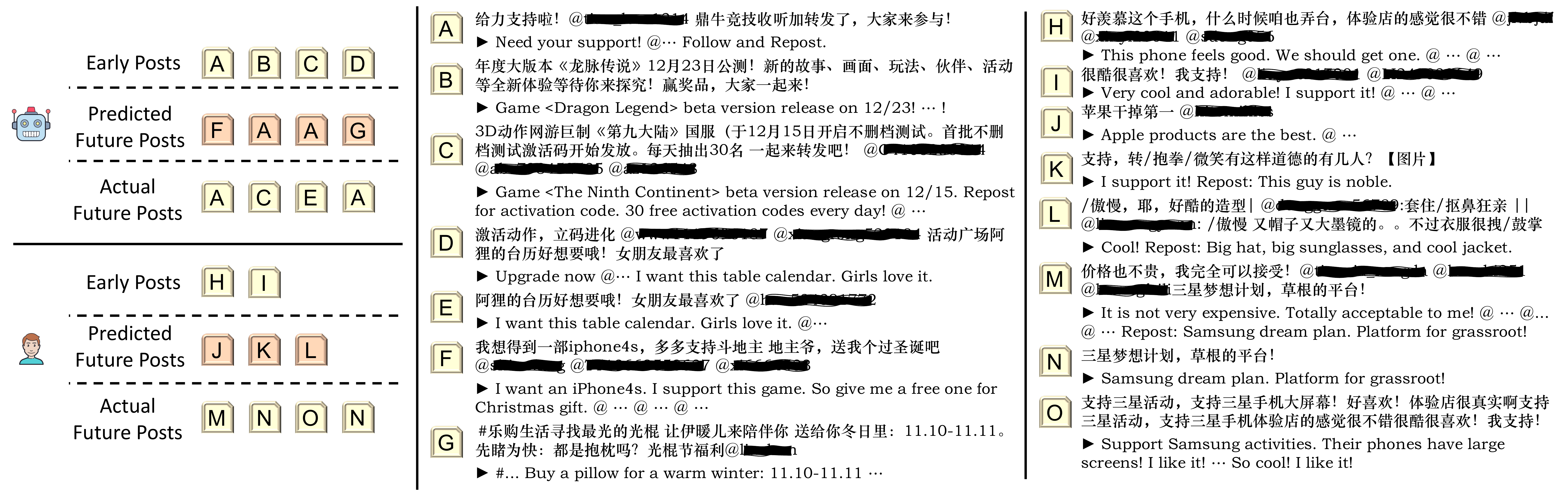}
\vspace{-0.1in}
\caption{Case study: the historical posts, predicted posts (by \methode), actual future posts sequences of two users, and actual post content each followed by a brief translation (marked with solid triangles). This study shows the previously wrong classified users can be correctly classified with the predicted posts by \methode.}
\label{fig:case_study}
% \vspace{-0.05in}
\end{figure*}

Particularly, for supervised methods, the performance of \methode with only 10\% or 20\% available data is better or comparable with the baselines with all data.
With only 20\% (\gcn) / 10\% (\gsage) / 10\% (\hetgnn) / 40\% (\dominant) / 60\% (\deepae) of earliest available data, \method achieves better or comparable performance than the original model with all available data.

Although the original performances are not monotonically increasing (though correlated), it is also worth mentioning that for all of the baseline detection methods, we are able to observe substantial improvements in absolute performance on average, which are evidence for \method's powerful ability to model the evolution of the graph structure and user-item interactions.

\subsubsection{Ablation study and sensitivity analysis}\hfill \\
\noindent \textbf{Ablation study on Seq2Seq choice.} 
To validate the robustness of \method on the choice of Seq2Seq model in the action sequence augmentation module, we show the results of \methode with different Seq2Seq models (GRU~\cite{cho2014learning}, LSTM~\cite{hochreiter1997long}, and traditional RNN~\cite{hochreiter1997long}) on Weibo dataset in Table~\ref{tab:ablation}. We can observe that GCN+\methode with all Seq2Seq methods can outperform vanilla GCN~\cite{kipf2016semi}, indicating that \method is robust to the choice of Seq2Seq method used in the action sequence augmentation module. Future works can try more powerful sequential models such as Transformer~\cite{vaswani2017attention} as the $g_{aug}$ module.

\vspace{0.03in}
\noindent \textbf{Hyperparameter sensitivity.} We test through \method's hyperparameters $\kappa$ and $\gamma$, which controls the amount of augmented actions for each user in \methodi and \methode, respectively. Figure~\ref{fig:sens} shows that the proposed \methodi and \methode are robust to $\kappa$ and $\gamma$ for range of $30 \le \{\kappa,\gamma\} \le 290$ for Weibo data.

\vspace{0.03in}
\noindent \textbf{Model convergence.}
Figure \ref{fig:conv} presents the convergence progress of \method. Figure \ref{fig:conv1} shows that the change of performance over iterations of \methodi with \gcn. We observe a smooth yet fast increase of the AUC at the first few iterations, which then reaches a steady state. The curve shows the bootstrapping iterative training design of \methodi is meaningful and demonstrates the mutual beneficial relationship between the two modules. Figure \ref{fig:conv2} shows the convergence of loss and validation AUC during training.

\vspace{-0.05in}
\subsection{Case Study}

To further demonstrate the effectiveness of our proposed \method framework, we conduct case studies on the Weibo data. We study two user cases: (1) an anomalous user who was falsely classified as a benign user at early stage when data were not complete, and (2) a benign user who was falsely classified as an anomaly at an early stage. Figure \ref{fig:case_study} presents their historical posts, predicted posts (given by \methode), and actual future posts. 

For the anomalous user, \gsage originally classified him as a benign user with confidence of 0.73. \methode successfully classified him as an anomaly with confidence of 0.58. From the posts of this user we observe that the user was posting advertisements in early posts. However, the posts did not repeat or show any strong pattern of an anomaly. \methode correctly predicted that he will repeat posting the message $A$ and hence enforced his suspicious pattern. The ground truth showed that this user actually did continue to repeat the advertisement posts (e.g., repeat posting message $A$). We verified that this user was an anomaly by checking the complete history of his posts and we found that the history has a large number of advertisements.

For the other user, \gsage falsely predicted this benign user as an anomaly with confidence of 0.73. With \methode, this user was correctly classified as a benign user with confidence of 0.74. One possible reason is that this user mentioned cell phones in his early posts and there were a large number of advertisement posts about phones in the dataset. \methode predicted that he would be posting some unrelated posts, one of which was related with iPhone. Although the predicted future posts were not exactly the same as the actual future posts, the predicted posts showed that the behavior pattern of this user is dissimilar to that of an anomalous user. We verified that this user was benign. Most of his posts were about personal ideas. His posts that supported Samsung phones caused the false positive classified by \gsage.

\section{Conclusions}
\label{conclusion}
In this work, we proposed \method for early graph-based anomaly detection via action sequence augmentation. \method aimed at improving existing graph anomaly detection methods with limited observations. Our work managed to model both the node representation and graph topology evolution through behavior forecasting via action sequence augmentation. Experiments on real-world data demonstrated that \method improves graph anomaly detection methods towards early and accurate anomaly detection.

\begin{acks}
This research was supported in part by Snap Research Fellowship and National Science Foundation (NSF) Grants no. IIS-1849816 and no. CCF-1901059.
\end{acks}

% \clearpage
\balance
\bibliographystyle{ACM-Reference-Format}
\bibliography{ref}

\end{document}